\pdfoutput=1

\documentclass[11pt]{article}

\usepackage[final]{latex/acl}

\usepackage{times}
\usepackage{latexsym}

\usepackage[T1]{fontenc}

\usepackage[utf8]{inputenc}

\usepackage{microtype}

\usepackage{inconsolata}

\usepackage{graphicx}
\usepackage{booktabs}

\usepackage{hyperref}
\usepackage[most]{tcolorbox}
\usepackage{array}
\usepackage{subcaption}
\usepackage{caption}
\usepackage{amsmath}
\usepackage{multirow}
\usepackage[nameinlink]{cleveref}
\usepackage{colortbl}
\usepackage{tabularx}

\usepackage{pifont}
\usepackage{xcolor}
\usepackage{amssymb}
\usepackage[most]{tcolorbox}
\usepackage{adjustbox}
\usepackage{roboto} 

\def\DatasetLong{{\textsc{BlockWorld-Repairs}}}
\def\DatasetShort{{BW-R}}

\newcommand{\emojiLetter}[2]{
\tcbox[enhanced,left=0.5mm,right=0.5mm,top=0.5mm,bottom=0.05mm,boxsep=-0.1mm,colframe=#1,colback=#1,nobeforeafter,fontupper=\roboto\bfseries\small]{#2}
}

\definecolor{softorange}{HTML}{fbe3d6}
\definecolor{pastelpurple}{HTML}{f2cfee}
\definecolor{softblue}{HTML}{caeefb}

\newcommand{\greentick}{\textcolor{green}{\ding{51}}}
\newcommand{\redcross}{\textcolor{red}{\ding{55}}}
\newcommand{\visualtype}{\emojiLetter{softorange}{V}}
\newcommand{\relationaltype}{\emojiLetter{pastelpurple}{R}}
\newcommand{\dialoguetype}{\emojiLetter{softblue}{D}}

\title{Repairs in a Block World: A New Benchmark for Handling User Corrections with Multi-Modal Language Models}

\author{
Javier Chiyah-Garcia ~~~ 
Alessandro Suglia ~~~ 
Arash Eshghi \\
  Heriot-Watt University, Edinburgh, United Kingdom \\
  \texttt{\{fjc3, a.suglia, a.eshghi\}@hw.ac.uk} \\
}

\begin{document}
\maketitle
\begin{abstract}

In dialogue, the addressee may initially misunderstand the speaker and respond erroneously, often prompting the speaker to correct the misunderstanding in the next turn with a Third Position Repair (TPR). The ability to process and respond appropriately to such repair sequences is thus crucial in conversational AI systems. In this paper, we first collect, analyse, and publicly release \DatasetLong: a dataset of multi-modal TPR sequences in an instruction-following manipulation task that is, by design, rife with referential ambiguity. We employ this dataset to evaluate several state-of-the-art Vision and Language Models (VLM) across multiple settings, focusing on their capability to process and accurately respond to TPRs and thus recover from miscommunication. We find that, compared to humans, all models significantly underperform in this task. We then show that VLMs can benefit from specialised losses targeting relevant tokens during fine-tuning, achieving better performance and generalising better to new scenarios. 
Our results suggest that these models are not yet ready to be deployed in multi-modal collaborative settings where repairs are common, and highlight the need to design training regimes and objectives that facilitate \textit{learning from interaction}. Our code and data are available at \url{www.github.com/JChiyah/blockworld-repairs}




\end{abstract}

\section{Introduction}
\begin{figure}[t]
\centering
\includegraphics[width=1\linewidth]{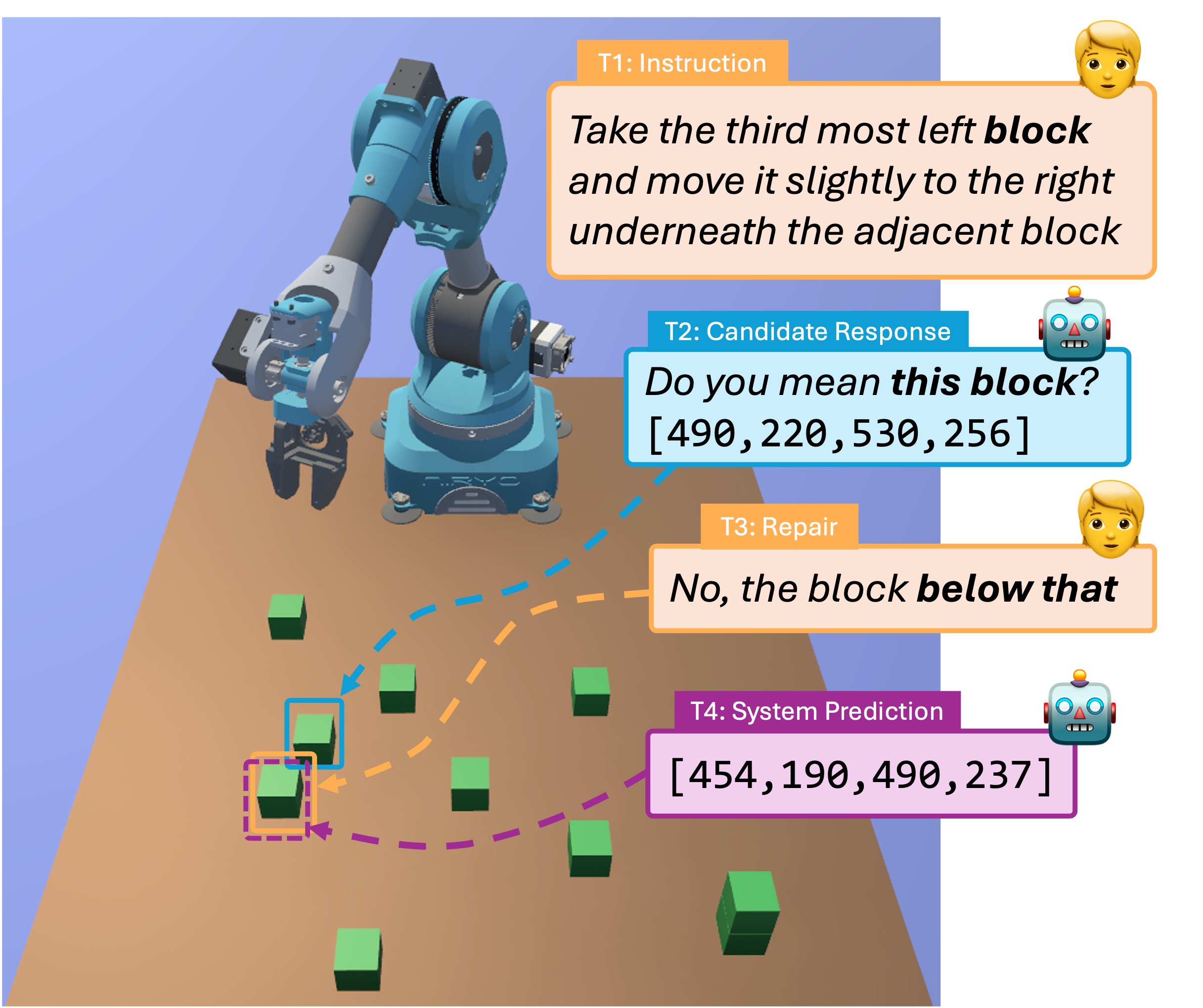}
\caption{Example dialogue from \DatasetLong: after predicting an incorrect response, VLMs must accurately interpret the repair to produce the correct bounding box prediction, a critical skill for human-robot collaboration tasks.}
\label{fig:sample_dialogue}
\end{figure}

Unlike its formulation in much of the literature within NLP \citep[see especially][]{glue,superglue}, Natural Language Understanding (NLU) is not a unilateral, passive process, but an (inter)active one \citep[see][for expansive discussion]{schlangen-2023-general}: in everyday conversation, people continuously work together to \textit{negotiate} shared understanding and coordination in order to move the conversation forward \cite{Clark96,Clark.Brennan91,Goodwin81,Healey.etal18,Mills07}. 
One of the key interactional processes that enables this is called \textit{repair} \cite{Schegloff.etal77,Schegloff92} -- see \Cref{fig:sample_dialogue}: a set of universal and highly systematised corrective feedback mechanisms for dealing with \textit{miscommunication} when it arises in conversation \cite{Enfield.etal13,Dingemanse.etal15}.

Therefore, the ability to interpret and generate effective repair sequences is crucial to \textit{robust} and \textit{faithful} Conversational AI technology. This need is especially acute in settings where more fine-grained levels of understanding are required, for example in embodied human-machine collaboration. Indeed the frequency of miscommunication in human dialogue is known to vary with both the overarching task and the medium of communication \cite{Colman.Healey11}.

In this paper, we focus on a particular class of corrective feedback, so called, \textit{Third Position Repairs} \citep[henceforth TPR;][] {Schegloff92} in multi-modal settings. These occur when the addressee initially misunderstands the speaker (see \Cref{fig:sample_dialogue} at T1, the \textit{trouble source} turn), responds based on this misunderstanding (at T2), which in turn reveals the misunderstanding to the addressee who then goes on to correct the misunderstanding (at T3). The addressee then provides a new response (at T4) to the original request (at T1) based on the correction. TPRs have been largely neglected in the NLP community, likely due to underestimation of the importance of repair phenomena in dialogue, and thereby lack of appropriate data and frameworks for model training and evaluation. Repair is of particular interest in (multi-modal) language modelling since it almost invariably involves highly focused / specific revisions to the representations already built by the model (e.g. revisions to the referents of referring expressions, like in our case here), based on the context-dependent structure of the corrective feedback. Repairs can thus be used as probes into how models represent meaning and the structure of the overarching task (see \newcite{chiyah-garcia-etal-2023-referring,madureira-schlangen-2023-instruction, benotti-blackburn-2021-recipe} for different implementations of this idea). And as we show here, this is also consequential for how language models should be trained to process TPRs including the specific objective functions involved. 

Our contributions are thus as follows: 1) we collect and release \DatasetLong{} (\DatasetShort), a dataset of collaborative dialogues in a tabletop manipulation task that, by design, has a high potential for referential ambiguity and is focused on complex multi-modal task instructions as well as TPRs; 2) we establish a human baseline for the whole task through an in-person human study where participants try to solve the task using these dialogues; 3) we show that state-of-the-art Vision Language Models (VLM) can learn to process TPR sequences through a specialised training regime, with some opportunity for generalisation; and 4) considering the substantial performance gap between the best models and humans, we present an in-depth error analysis comparing the two, revealing that models struggle with references that humans find easy.

\section{Background}

\paragraph{Communicative Grounding, Miscommunication and Repair} What is often ignored or glossed over in today’s research on Conversational AI is that conversation involves collaborative effort from speakers and addressees to ensure that what is said is understood sufficiently for the task at hand before the conversation can move forward: this almost continuous exchange of both positive and negative feedback is called \textit{communicative grounding}\footnote{Not to be confused with \textit{symbol grounding} but see \cite{Larsson18} for how the two are related.} \citep[see][and many others following]{Clark96}. This exchange of feedback enables interactants to coordinate their (linguistic and non-linguistic) actions in conversation \cite{Eshghi.etal23,Mills14}, and enables shared languages to be established and sustained \cite{Healey.etal18}. \textit{Miscommunication} occurs when one of the interactants detects a problem in their own, or another's understanding, and is usually dealt with immediately using different forms of corrective feedback, or, as they are collectively called, \textit{repair} \cite{Schegloff92}. These are classified in the literature based on who initiates the repair; who performs it; and where the actual correction takes place. For example, Clarificational Exchanges (CE) whereby the addressee produces a request for clarification (CR) and the original speaker provides a response / correction in the next turn, constitute other-initiated, self-repairs. On the other hand, TPRs, our focus in this paper, are self-initiated, self-repairs on the third turn.

\paragraph{Computational Models of Repair} Considerable attention has been paid to computational models for the interpretation and generation of \textit{self-repair} \citep[see][among others]{Hough.Schlangen15,Hough15,Shalyminov.etal17,Eshghi.Ashrafzadeh23,Buss.Schlangen11,Hough.Purver12}: a class of repairs whereby the speaker corrects themselves on the fly within the same conversational turn (e.g.\ ``User: I want to go to London uhm sorry Paris''). Similarly, the crucial role of generating and responding to CRs (e.g.\ ``Pardon/what/who?'') in conversational models has long been recognised \citep[see][among others]{San-Segundo.etal01, Purver04a,Purver.Ginzburg04,Rieser.Moore05,Rodriguez.Schlangen04,Rieser.Lemon06}, but existing systems either remain limited (e.g.\ \newcite{Curry.etal18}) or do not support this at all -- see \newcite{Purver.etal18} for an overview. Recent work tries to identify when to pose a CR \cite{addlesee-etal-2024-clarifying, madureira-schlangen-2024-taking, madureira-schlangen-2023-instruction, zhu-2021, shi-etal-2022-learning, addlesee-eshghi-2021-incremental}, but few evaluate the ability of models to process their responses \cite{Gervits.etal21, aliannejadi-etal-2021-building}. \newcite{chiyah-garcia-etal-2023-referring} show that the ability of Vision and Language Models (VLM) to effectively process clarificational exchanges depends on the level of granularity of the model’s cross-modal representation, and therefore also the specific objectives used to train the models.

Our work here is closest in spirit to that of \newcite{balaraman-etal-2023-thats} who provide the first large dataset of Third Position Repairs, and use it to evaluate LLMs. However, they focus only on unimodal repairs in the context of Conversational Question Answering, and do not perform any fine-tuning of the models they evaluate, like we do here.

\paragraph{VLMs for Situated Collaborative Tasks}

Thanks to recent advances in the development of LLMs, there have been many attempts to derive Vision+Language models that use pretrained visual encoders to solve vision-language tasks (e.g., \cite{liu2023llava, laurençon2024matters}, \textit{inter alia}). It is important to note that most of these models are trained to follow instructions that can be specified in a single turn. Many tasks can be included in this category such as visual question answering~\cite{antol2015vqa}, image captioning~\cite{lin2014coco}, and referential expression resolution~\cite{yu2016refcoco}. To solve these tasks, models are trained to maximise the likelihood of the responses in the training data using supervised learning methods or using reinforcement learning from human feedback~\cite{ouyang2022training}, ignoring the fact that a conversation is a process that unfolds over multiple turns where each of which matters to make correct decisions. For instance, in tasks such as Visual Dialogue~\cite{das2017visual}, it is well-known that a very limited number of responses are dependent on the dialogue history~\cite{agarwal-etal-2020-history}, making them not a suitable benchmark for truly collaborative tasks requiring the ability to establish common ground. As highlighted by \citet{suglia2024visually}, many Embodied AI benchmarks (e.g., ALFRED~\cite{shridhar2020alfred}, Simbot Arena~\cite{gao2023alexaarenausercentricinteractive}, etc.) also have similar issues considering that the level of ambiguity is reduced to the minimum, minimising the need for any form of correction. 
For this reason, \DatasetShort{} represents the first benchmark that is aimed at assessing the ability of current VLMs to resolve TPRs---an important capability which is essential for establishing common ground.

\section{The BlockWorld-Repair Dataset}

We build on top of the Block World dataset by \citep{bisk-etal-2016-natural}, an instruction-following task where a robot manipulator has to move blocks on a virtual board. 
Humans provided instructions in natural language referring to a block to pick up and a new location to drop it with complex visual and spatial descriptions (i.e., \textit{``the block on the top-right corner moves behind the middle-top block''}). 

Prior works have struggled to handle the most ambiguous setting with blank blocks \cite{tan2018source, mehta-goldwasser-2019-improving, dan-etal-2021-generalization}.
However, here we argue that 
single-turn instructions are insufficient to solve highly ambiguous environments or real-world situations. Even the most detailed referring expressions in the Block World may fail to uniquely identify a referent (see \Cref{fig:sample_dialogue}), which repairs could alleviate by narrowing down the candidate pool and introducing new information to identify the correct referent (T3 in \Cref{fig:sample_dialogue}).
Similar to how humans misinterpret and subsequently repair referential ambiguities with TPRs, 
VLMs must also be capable of handling multi-modal TPRs in dialogues, as they are essential for mutual coordination. 





\paragraph{Differences from the original Block World}
The original dataset consists of pairs of images and single-turn instructions, requiring agents to resolve inherently ambiguous commands. However, it does not account for dialogue or clarificational exchanges as strategies to overcome environmental ambiguity, which humans commonly use to repair miscommunications. To address this, we extend the dataset by incorporating brief dialogues where humans interact with the agent and produce Third Position Repairs. We preserve the original Block World instructions and test set while adding our dataset as an extension to explore how humans resolve ambiguity through clarification and repair strategies.


\paragraph{Ecological validity} In contrast to more recent photo-realistic simulated environments (e.g., AI2Thor~\cite{kolve2017ai2}), we build \DatasetShort{} using the Unity engine to simulate a realistic tabletop manipulation task that gives us the ability to specifically control for the conditions that require collaboration between agents to correctly complete the task. This is also similar to previous works that sacrifice photo-realistic vision to asses the systematic generalisation ability of VLMs for tabletop manipulation tasks (e.g., VIMA-Bench \cite{pmlr-v202-jiang23b}), or works that focus on dialogue coordination in highly ambiguous scenarios (e.g., The Cups dataset \cite{Dobnik2020}). \Cref{ap:related_work} provides further comparisons to related datasets.


\subsection{Dataset Collection}

We built a dialogue interface on Amazon Mechanical Turk (AMT) where workers chat with a robot agent and provide instructions to move blocks in a virtual world, 
similar to the original dataset.
The robot agent is 
prone to mistakes and verifies its actions by pointing to blocks or locations. AMT workers can then collaborate with the robot to correct misunderstandings via a short dialogue through
complex, context-dependent TPRs, based on the robot's indicated position. 
\Cref{fig:sample_dialogue} shows an example dialogue.

The robot performs the correct action 70\% of the time\footnote{For comparison, the best models only pick the correct block 54\% of the time \cite{tan2018source}.} and selects the appropriate block or location after the corresponding repair, irrespective of the quality of the TPR.
Consequently,
humans provided at most one initial instruction specifying the block to pick and where to place it, one TPR indicating the \textbf{source block} to move, and one TPR specifying the \textbf{target position} for the block. The initial instructions are thus equivalent to the original commands. 





As a result of this data collection, we obtained 795 dialogues after filtering out low-quality and problematic ones, which include 795 initial instructions, 629 source block TPRs and 635 final target position TPRs, resulting in 2059 total entries. 
We provide further details about the AMT setup, task and dataset statistics in \Cref{ap:dataset_collection}.

\subsection{Quality Control and Human Baselines}


Since we used an automated agent to collect the dialogues and AMT is prone to provide low-quality data \cite{Saravanos2021}, we validated a data subset through an in-person study. We recruited 22 participants to take on the robot agent's role, following the instructions and repairs previously given by human workers.


Participants interacted with a similar interface to the data collection, with a randomly selected dialogue displayed in the chat window. Their task was to choose the source block and a target position they believed the dialogue referenced within a large image of the virtual table. They did not have access to the true block configurations and received no feedback on whether they selected the correct block/location. 
Participants did not see repeated entries (same dialogue), and at least two participants annotated each entry. 
After filtering out incorrect or problematic entries, we obtained 991 action annotations. 

Participants performed well overall and rated the AMT instructions highly, achieving 68\% accuracy in selecting the correct block to move, rising to 75\% after repairs, with similar success in target position prediction. These results emphasise the need for models to process TPRs accurately. We analyse human performance in §\ref{sec:error}.




\section{Experiments}

This section evaluates the ability of VLMs to process instructions and repairs in an instruction-following task within a situated context. The repair turns in the \DatasetShort{} are meaningful only when interpreted alongside the robot's pointing position 
at the time of the repair, based on the first instruction; thus, they are incomplete on their own. Consequently, we have dialogue triplets that are intrinsically connected and can only be comprehended as a whole: the \textit{initial instruction}, the \textit{incorrect candidate prediction}, and the \textit{repair}. We argue that more general models must be able to process both initial instructions and any subsequent repairs.




\subsection{Experimental Setup}

\subsubsection{Dataset}

We combine the Block World and \DatasetShort data (70/30 train/test), which contains two types of entries: \textit{single-instructions} (1 turn) and \textit{TPRs} (3 turns: instruction, candidate response, repair). We ablate the data during our experiments to analyse fine-grained information about the models' capabilities. Refer to \Cref{ap:dataset} for further dataset details. 


\subsubsection{Tasks}

Inspired by \citet{bisk-etal-2016-natural}, we evaluate models on two prediction tasks: 1) \textit{Source block prediction}, which consists of predicting the block to pick up out of the 10 candidates in the virtual world; and 
2) \textit{Target position prediction}, which consists of predicting where to drop the block on the table.

Previous works have formulated these tasks to aid the prediction, such as finding a reference block and an offset distance from it \cite{bisk-etal-2016-natural} or quadrants in a board \cite{mehta-goldwasser-2019-improving}. 
For our experiments, and following common visual grounding objectives in VLMs \cite{wang2022ofa,liu2023llava,chen2023shikra}, we prompt models to generate the coordinates of the bounding box for the block or location ($[x_{min}, y_{min}, x_{max}, y_{max}]$) in the range of 0-1, which we then scale to the image size. Predictions not matching this format during parsing are marked as out of distribution and removed. We use specific prompts for each model to match their training prompting strategy (see \Cref{ap:experimental_setup}).



\subsubsection{Metrics}

For evaluation, we use the metrics originally proposed in the Block World~\cite{bisk-etal-2016-natural}. For \textbf{source block accuracy}, we use Intersection over Union and select blocks above a threshold. We compute \textbf{block distances} by converting bounding boxes to their respective XYZ coordinates in the virtual world and then calculating the distance in blocks between the predicted and true locations.





\subsubsection{Models}\label{sec:models}

We experiment with two state-of-the-art open-source models: Idefics2 8B \cite{laurençon2024matters}, and LLaVA 1.5 7B \cite{liu2023improvedllava}. Both models have strong visual-textual cross-modality grounding capabilities and have been pre-trained on a mix of relevant vision and language objectives which facilitate bounding box generation. 
After preliminary tests with the ``chatty'' version of Idefics2, we decided to use the instruction-tuned version in all our experiments.
We also provide zero-shot evaluations for GPT-4o \cite{gpt4o}, a frontier VLM model~\footnote{To facilitate further analysis, we will release its predictions as part of our code release.}.

\subsection{Evaluating Models on TPRs}\label{sec:model_results}

VLMs should inherently be able to process TPRs as they are fundamental components of human dialogues. We initially assess the VLMs' 
out-of-the-box capabilities in a zero-shot setting, and then proceed to distil these capabilities through fine-tuning. In both cases, we prompt the models with the task instruction, followed by the conversation and provide the relevant image of the blocks. We fine-tune for 1 epoch using parameter-efficient methods (e.g., LoRA~\cite{hu2021lora}) following the recommended hyperparameters for each model. We train a different model for each sub-task so models can fully leverage the input prompt and avoid interference between the two tasks (i.e., source block and target position prediction). 


\begin{table}[ht]
    \centering
    \addtolength{\tabcolsep}{-0.4em}
    \small{
\begin{tabular}{@{}cll|cc|c@{}}
    \toprule
    \multirow{1}{*}{\textbf{Test}} & \textbf{Train} & \multirow{2}{*}{\textbf{Model}} & \multicolumn{2}{c|}{\textbf{Source}}                   & \textbf{Target}             \\
    \multirow{1}{*}{\textbf{Data}} & \textbf{Data} &                                 & \textbf{Acc ↑} & \textbf{Mean (SD) ↓} & \textbf{Mean (SD) ↓} \\
    \midrule

 &              & Idefics2-zs   & 0.18 & 5.06 (±3.5) & 9.55 (±2.5) \\
 &              & LLaVA-zs & 0.24 & 4.40 (±3.3) & 5.74 (±3.0) \\
 \rowcolor[HTML]{EFEFEF} \cellcolor{white}
 & IO & Idefics2-ft   & 0.33 & 3.33 (±3.0) & 4.93 (±2.6) \\
 \rowcolor[HTML]{EFEFEF} \cellcolor{white}
 & IO & LLaVA-ft & 0.25 & 3.77 (±2.9) & 5.06 (±2.9) \\
 & RO  & Idefics2-ft   & 0.21 & 4.36 (±3.1) & 5.38 (±2.4) \\
 & RO  & LLaVA-ft & 0.12 & 5.05 (±3.1) & 6.32 (±2.9) \\
 \rowcolor[HTML]{EFEFEF} \cellcolor{white}
 & Full         & Idefics2-ft   & 0.30 & 3.39 (±2.9) & 4.75 (±2.6) \\
 \rowcolor[HTML]{EFEFEF} \cellcolor{white}
 & Full         & LLaVA-ft & 0.22 & 3.82 (±2.9) & 4.46 (±2.4) \\
 \multirow{-9}{*}{\rotatebox[origin=c]{90}{\texttt{Instructions}}}
 &              & GPT-4o-zs        & 0.30 & 3.83 (±3.3) & 4.22 (±2.6) \\
 
 \midrule

 &              & Idefics2-zs   & 0.00 & 3.74 (±1.9) & 4.13 (±2.1) \\
 &              & LLaVA-zs & 0.13 & 3.29 (±2.1) & 3.45 (±1.2) \\
 \rowcolor[HTML]{EFEFEF} \cellcolor{white}
 & IO & Idefics2-ft   & 0.21 & 4.19 (±2.7) & 5.11 (±2.6) \\
 \rowcolor[HTML]{EFEFEF} \cellcolor{white}
 & IO & LLaVA-ft & 0.18 & 3.43 (±2.4) & 3.49 (±1.3) \\
 & RO  & Idefics2-ft   & 0.58 & 2.19 (±3.1) & 7.07 (±2.7) \\
 & RO  & LLaVA-ft & 0.07 & 3.20 (±1.7) & 3.67 (±1.3) \\
 \rowcolor[HTML]{EFEFEF} \cellcolor{white}
 & Full         & Idefics2-ft   & 0.26 & 3.43 (±2.4) & 5.81 (±2.3) \\
 \rowcolor[HTML]{EFEFEF} \cellcolor{white}
 & Full         & LLaVA-ft & 0.44 & 2.66 (±2.8) & 3.69 (±2.2) \\
\multirow{-9}{*}{\rotatebox[origin=c]{90}{\texttt{Repairs}}}
 &              & GPT-4o-zs        & 0.41 & 2.75 (±3.1) & 2.95 (±2.2) \\
\midrule
& \multicolumn{2}{c|}{Random Baseline}  & 0.10 & 6.50 (±3.0) & 6.26 (±2.9) \\
\bottomrule
    \end{tabular}}
    \caption{Model performance on source selection and target position prediction tasks on zero-shot (zs) or fine-tuned (ft) on data subsets: Instructions-Only (IO), Repairs-Only (RO) and Full. We compare source block accuracy (↑) and mean block distances (↓). Lower distances indicate predictions closer to the correct location.}
    \label{tab:model_performance}
\end{table}

\paragraph{Results} \Cref{tab:model_performance} shows the performance of testing the VLMs on either instruction or repair-only data subsets with a mix of training \DatasetShort{} data.
Zero-shot models typically perform above the random baseline, initially demonstrating their visual-grounding capabilities. Fine-tuned models exhibit more nuanced results: they partially learn the task through training, particularly when trained and tested on the same type of entries (instructions or repairs). However, employing the full dataset does not consistently enhance performance: Idefics2 appears to adapt well to the training data or shows greater data efficiency, whereas LLaVA achieves better results with the full data and therefore suggests stronger generalisation capabilities. 

Overall, the models struggle to exploit TPRs, especially in zero-shot settings. Intuitively, repairs should facilitate the task by introducing new information that refines the candidate pool. However, because repairs are often context-dependent (i.e., based on the candidate response), models require strong object-centric representations \cite{bengio2013representation, Seitzer2023BridgingTheGap} and a deeper understanding of relational dynamics between objects to effectively leverage TPRs (e.g., ``the block to the right'') \cite{Ilinykh2021}. This is a particular issue of current vision transformers, which frequently struggle with positional concepts such as spatial understanding \cite{ilinykh-dobnik-2022-attention, pantazopoulos-etal-2024-lost}, and tend to rely on language biases \cite{Salin2022}.
GPT-4o is unique in its ability to process these repairs effectively and improve its performance with instructions. We see that models can learn to process TPRs with some training, and the following section explores methods to further distil this capability.

\section{Learning to Process TPRs}\label{sec:learning_tprs}

During training, models encountered two types of entries: single-turn instructions or dialogues up to the TPR. In human conversations, we can effortlessly manage both scenarios; therefore it is reasonable to expect that models trained to predict bounding boxes from a single instruction should also be capable of handling repairs to their predictions. 
However, our previous results indicate that achieving both capabilities simultaneously is challenging, and that generalization incurs a cost, even when models address the same task.

To handle repairs, models cannot rely only on the last turn but must consider the context from the initial user instruction and the following candidate output. The intermediate predicted bounding box, even if inaccurate, also provides crucial information within the context of the TPR, as repairs rarely repeat a fully-formed instruction (i.e., \textit{``The right one!''}). In the case of \DatasetShort, repairs are relative to where the robot is currently pointing and often provide partial information to correct the robot's action. During fine-tuning, following common approaches in VLM training~\citep{laurençon2024matters}, VLMs calculate the cross-entropy loss for all the tokens in the input including both the intermediate (wrong) and last (correct) generation.
Thus, the models from §\ref{sec:model_results} are learning from incorrect tokens, and therefore these tokens are somehow affecting the quality of the VLMs generations. 





\subsection{Masking Token Loss}

\begin{figure}[ht]
    \centering
    \includegraphics[width=1\linewidth]{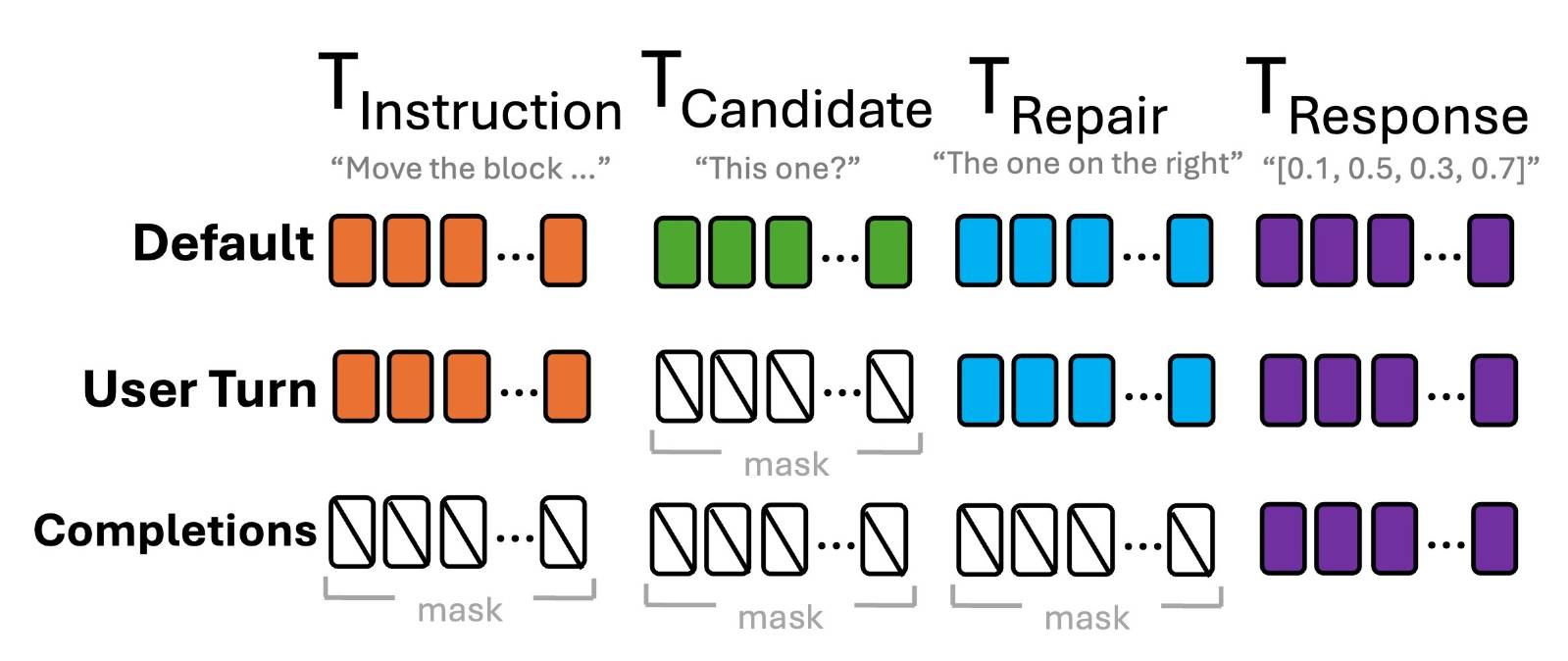} 
    \caption{Masking criteria for the cross-entropy loss.}
    \label{fig:masking_token_loss}
\end{figure}

To encourage VLMs to process and correctly handle TPRs, we experiment with different training regimes during fine-tuning. Specifically, as shown in \Cref{fig:masking_token_loss}, we define different masking criteria for the cross-entropy loss and assess their effect on downstream performance. 


\noindent \textbf{Default loss}, VLM's default cross-entropy applied to all tokens~\cite{laurençon2024matters}.

\noindent \textbf{User-turn loss}, we only calculate the loss for the user turns and the completion target. In this case, the intermediary prediction would not influence the overall loss but models could still benefit from learning instructions and the task format.

\noindent \textbf{Completion-only loss} to focus solely on the generated tokens and ignore all previous user and assistant turns in the entry. We expect these models to have the strongest generalisation capabilities, at the cost of not learning about the input instructions or corrections. 

 We follow the same procedure for fine-tuning as in \Cref{sec:model_results}, training one model per sub-task with the default hyperparameters.


\subsection{Results}

\Cref{tab:masking_summary} shows the fine-tuned VLMs on the full data across different masking criteria\footnote{All model results are provided in \Cref{ap:additional_results}.}. We first observe that applying a mask on all tokens but completions has large benefits for both VLMs for repairs in source prediction, with strong improvements particularly when fine-tuning with repairs-only or the full dataset. This loss allows models to learn from TPRs and transfer some task knowledge from instructions to repairs when using the full data. 
In other settings, this loss has small negative effects on the VLM's performance, especially when models are trained and tested on the same data. This is somewhat expected as calculating the loss for the complete input has benefits, such as using many more tokens during training, learning the input format and quickly adapting to the task. However, these models do not seem to learn the task well and thus do not generalise. 

\begin{table}[htb]
\centering
\addtolength{\tabcolsep}{-0.4em}
\small{

\begin{tabular}{@{}cll|cc|c@{}}
\toprule
& \multirow{2}{*}{\textbf{Model}}                          & \multirow{2}{*}{\textbf{Loss}} & \multicolumn{2}{c|}{\textbf{Source}}            & \textbf{Target}      \\
                       &            &               & \textbf{Acc ↑}  & \textbf{Mean (SD) ↓} & \textbf{Mean (SD) ↓} \\
\midrule

& \multirow{3}{*}{Idefics2} & Default       & 0.30            & 3.39 (±2.9)          & 4.75 (±2.6)          \\
&                                   & User-turn     & 0.36            & 3.12 (±3.0)          & 4.59 (±2.6)          \\
&                                   & Completion    & 0.33            & 3.33 (±3.0)          & 4.52 (±2.6)          \\
\rowcolor[HTML]{EFEFEF} \cellcolor{white}
                        &            & Default       & 0.22            & 3.82 (±2.9)          & 4.46 (±2.5)          \\
\rowcolor[HTML]{EFEFEF} \cellcolor{white}
&                                   & User-turn     & 0.24            & 3.65 (±2.9)          & 4.60 (±2.5)          \\
\rowcolor[HTML]{EFEFEF} \cellcolor{white}
\multirow{-6}{*}{\rotatebox[origin=c]{90}{\texttt{Instructions}}} &  \multirow{-3}{*}{LLaVA}                                 & Completion    & 0.19            & 3.93 (±2.7)          & 4.51 (±2.8)          \\

\midrule
                        & \multirow{3}{*}{Idefics2}             & Default       & 0.26            & 3.43 (±2.4)          & 5.81 (±2.4)          \\
&                                   & User-turn     & 0.32            & 2.94 (±2.6)          & 6.00 (±2.0)          \\
&                                   & Completion    & 0.47            & 2.29 (±2.4)          & 5.90 (±2.7)          \\
 \rowcolor[HTML]{EFEFEF} \cellcolor{white}
                        &            & Default       & 0.44            & 2.66 (±2.8)          & 3.69 (±2.2)          \\
 \rowcolor[HTML]{EFEFEF} \cellcolor{white}
&   \multirow{3}{*}{LLaVA}                                & User-turn     & 0.37            & 2.69 (±2.5)          & 4.04 (±2.4)          \\
 \rowcolor[HTML]{EFEFEF} \cellcolor{white}
\multirow{-6}{*}{\rotatebox[origin=c]{90}{\texttt{Repairs}}}
 &  \multirow{-3}{*}{LLaVA}                                 & Completion    & 0.54            & 2.01 (±2.5)          & 4.56 (±2.7)         \\

\bottomrule
\end{tabular}}
\caption{Fine-tuned VLMs with different loss criteria.}
\label{tab:masking_summary}
\end{table}

One of our assumptions is that the intermediate predictions hurt models, as these would calculate losses for incorrect tokens. The results of masking these tokens in the loss (user-turns only) suggest that this loss does not help the VLMs as much as completion loss, and has a mixed bag of effects (positive or negative). 

Most of the improvements with either of the losses only apply to source prediction. The models struggle with target position predictions and do not show clear benefits from these losses. Although these sub-tasks expect a similar output, predicting a target location is much harder, as it does not depend on a particular salient object in the image (i.e., a block) but on spaces between objects. It is thus unsurprising that VLMs pre-trained with visual grounding objectives (e.g., RefCOCO \cite{yu2016refcoco}) do not perform well when we are not referring to an object. In these cases, it seems that the additional prompt tokens are helpful for models to learn the task and masking them hurts their performance. 




Furthermore, we see that models fine-tuned on instructions alone, with or without masking, acquire some task capabilities, but do not generalise well to repairs, and vice-versa. When trained with the full data and completion-only loss, VLMs learn what is important regardless of the conversation, and reach their best results, learning what is important particularly for repairs. 


During these fine-tuning experiments, we identify two main factors at play: 1) the size of the training data; and 2) the incorrect candidate responses. When training on repairs, the size (i) becomes the bigger issue as masking considerably reduces the available tokens during training, which is crucial when data is smaller. Models do better when there is more training data, as happens with instructions, but they train on incorrect bounding boxes from the intermediate system turn (ii), hurting their performance. Handling repairs allows us to evaluate the models' ability to interpret the initial instruction, candidate response, and repair turns. Calculating losses on completions only ensures that models learn a more robust prompt representation that is useful when generalising across both types of data.




\section{Error Analysis}\label{sec:error}

We want models that can collaborate with humans even when misunderstandings arise, and thus we need to understand the differences in how models and humans solve the same visual-grounding task. This section attempts to analyse models beyond performance metrics and provides insight into the models' behaviour, such as where mismatches arise compared to how humans process TPRs. We use the best models from §\ref{sec:learning_tprs} trained on the whole data using completions-only loss.


\subsection{Human Comparison}

\begin{table}[t]
\centering
\addtolength{\tabcolsep}{-0.35em}
\small{
\begin{tabular}{@{}cl|cc|c@{}}
\toprule
\multicolumn{1}{l}{\textbf{}}          & \multirow{2}{*}{\textbf{Model}}         & \multicolumn{2}{c}{\textbf{Source}}     & \textbf{Target} \\
\multicolumn{1}{l}{}                   & \textbf{}              & \textbf{Acc ↑} & \textbf{Mean (SD) ↓} & \textbf{Mean (SD) ↓} \\
\midrule

\multirow{4}{*}{\rotatebox[origin=c]{90}{\texttt{Inst.}}} & Idefics2 - ft   & 0.21                & 3.97 (±4.0)     & 5.52 (±3.1)     \\
                                       & LLaVA - ft & 0.16                & 4.08 (±4.1)     & 4.63 (±3.4)     \\
                                       & GPT-4o        & 0.26                & 4.60 (±4.6)     & 4.30 (±3.4)     \\
                                       & Human Participants      & 0.68                & 1.59 (±1.6)     & 3.64 (±3.4)     \\
               \midrule
\multirow{4}{*}{\rotatebox[origin=c]{90}{\texttt{Repairs}}}  & Idefics2 - ft   & 0.43                & 2.46 (±2.5)     & 5.80 (±2.7)     \\
                                       & LLaVA - ft & 0.60                & 1.82 (±1.8)     & 4.82 (±2.7)     \\
                                       & GPT-4o        & 0.50                & 2.78 (±2.8)     & 2.58 (±1.8)     \\
                                       & Human Participants      & 0.75                & 1.41 (±1.4)     & 2.77 (±2.8)    \\
 
 \bottomrule
\end{tabular}}
\caption{Humans compared to the best models for the \DatasetShort~test subset with human annotations.}
\label{tab:human_performance}
\end{table}

To compare models with humans on this task, we use the data samples that we collected from our in-person human study, in which participants saw the same data that models are evaluated on.
The results in \Cref{tab:human_performance} show clear differences, with models struggling to reach human performance but demonstrating significant benefits from TPRs after training. We can see that, despite Idefics2 and LLaVA having lower source accuracy for instructions, they both beat GPT-4o on distance for source. In this case, accuracy fails as a metric for prediction quality, since lower distances better reflect true proximity. We also see impressive results for LLaVA processing TPRs after our training for source prediction. Regarding target position, this task is considerably more challenging for humans and models, reflected in worse scores overall. 
LLaVA and Idefics2 struggle to process repairs related to target locations, in contrast to GPT-4o which outperforms the human participants. 



\subsection{Impact of Task Difficulty}

To better pinpoint the strengths and weaknesses of the different models, we categorise the \DatasetShort~data into difficulty levels according to human performance\footnote{We provide this analysis according to GPT-4o performance in \Cref{ap:error_analysis}.}. We use lower human performance as an indirect proxy for the complexity of the example, and define three levels as follows: 1) \textbf{Easy:} both human annotators for the same entry correctly predicted the block (100\% source accuracy) or their target position was within 1 block unit distance away from the true location; 2) \textbf{Medium:} at least one out of the two annotators correctly predicted the block (50\% source accuracy) or their target position was below the human's mean distance away from the location (between 1 and 3.22 units away); 3) \textbf{Hard:} neither of the annotators correctly predicted the block (0\% source accuracy) or their target position was further than the human's mean distance (above 3.22 units).

\begin{table}[t]
\centering
\addtolength{\tabcolsep}{-0.3em}
\footnotesize{
\begin{tabular}{@{}ll|cc|c@{}}
\toprule
\multirow{2}{*}{\textbf{Difficulty}} & \multirow{2}{*}{\textbf{Model}}         & \multicolumn{2}{c}{\textbf{Source}}     & \textbf{Target} \\
\textbf{}          &      & \textbf{Acc ↑} & \textbf{Mean (SD) ↓} & \textbf{Mean (SD) ↓} \\
\midrule

Easy & Idefics2               & 0.35 & 3.09 (±3.1) & 5.72 (±3.1) \\
& LLaVA             & 0.41 & 2.62 (±2.6) & 3.98 (±2.5) \\
& GPT-4o                    & 0.45 & 3.50 (±3.5) & 2.18 (±1.3) \\
& Humans   & 1.00 & .00 (±.0)   & .81 (±.5)   \\
\midrule
Medium & Idefics2               & 0.37 & 2.32 (±2.3) & 5.30 (±2.3) \\
& LLaVA             & 0.42 & 2.19 (±2.2) & 4.54 (±3.0) \\
& GPT-4o                    & 0.42 & 3.05 (±3.1) & 2.86 (±2.0) \\
& Humans & 0.50 & 2.24 (±2.2) & 2.28 (±1.3) \\
\midrule
Hard & Idefics2               & 0.22 & 3.82 (±3.8) & 6.00 (±3.2) \\
& LLaVA             & 0.25 & 4.30 (±4.3) & 5.77 (±3.3) \\
& GPT-4o                    & 0.16 & 4.46 (±4.5) & 5.30 (±3.7) \\
& Humans   & 0.00 & 5.62 (±5.6) & 6.25 (±3.3) \\

\bottomrule
\end{tabular}}
\caption{Model performances across subsets with increasing levels of complexity (\textit{easy, medium, hard}).}
\label{tab:difficulty_levels}
\end{table}

The results of evaluating VLMs on these difficulty subsets (\Cref{tab:difficulty_levels}) indicate that models and humans do not struggle on the same entries. Models show a more uniform distribution of performance, with improvements over humans in the most difficult subset but significantly underperforming with easier references. This behaviour seems to follow the trend with the average word number in the subsets\footnote{See \Cref{ap:error_analysis}.}, suggesting that easy examples (most words) may employ longer, more complex terms to describe references, whereas hard examples (fewest words) may have easier terms for models, but are too underspecified for humans.


\subsection{Qualitative Analysis}

\begin{figure}[t]
    \centering
    \begin{subfigure}{\linewidth}
        \centering
        \caption{Source block selection}
        \includegraphics[width=\linewidth]{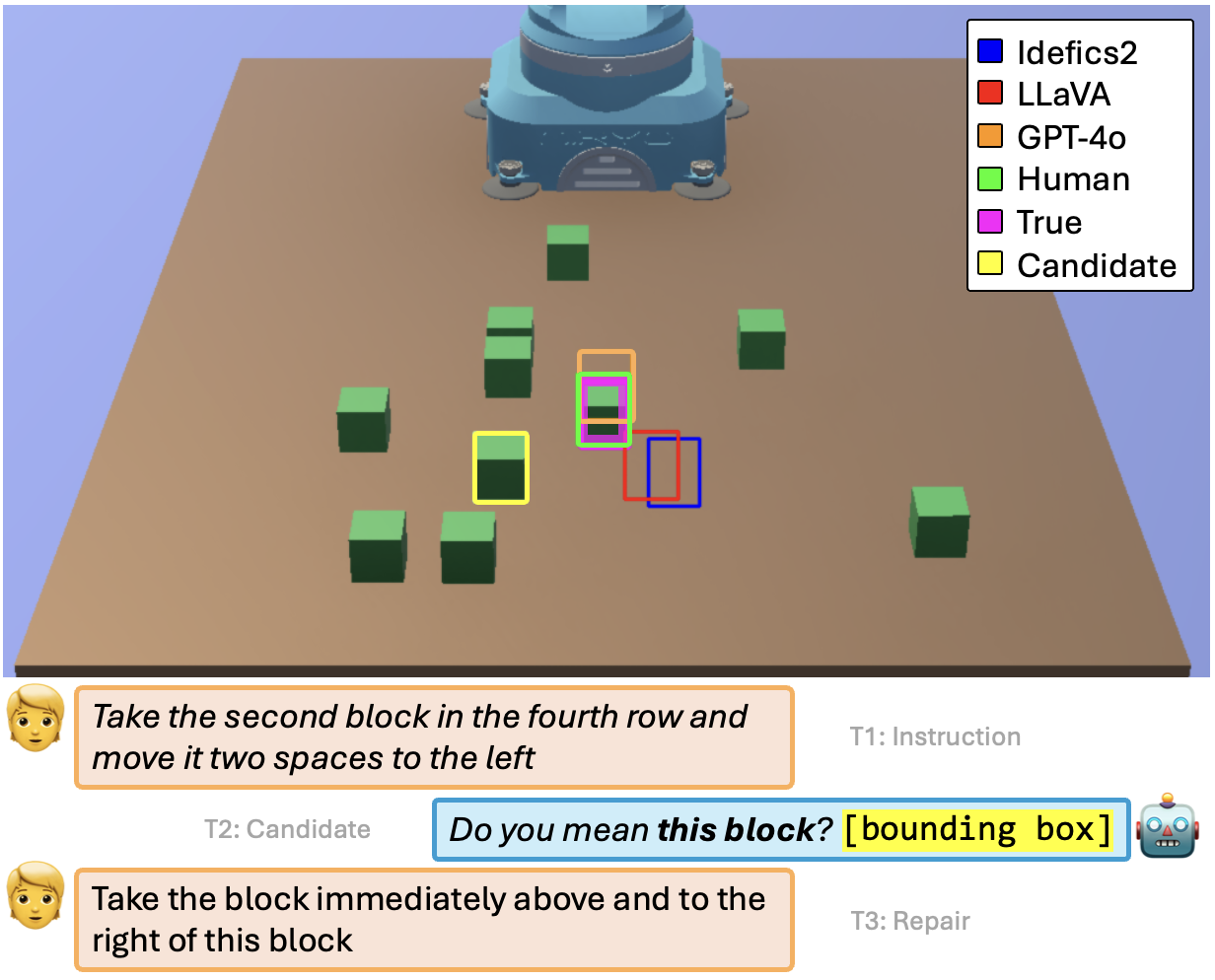}
    \end{subfigure}
    

    \begin{subfigure}{\linewidth}
        \centering
        \caption{Target position prediction}
        \includegraphics[width=\linewidth]{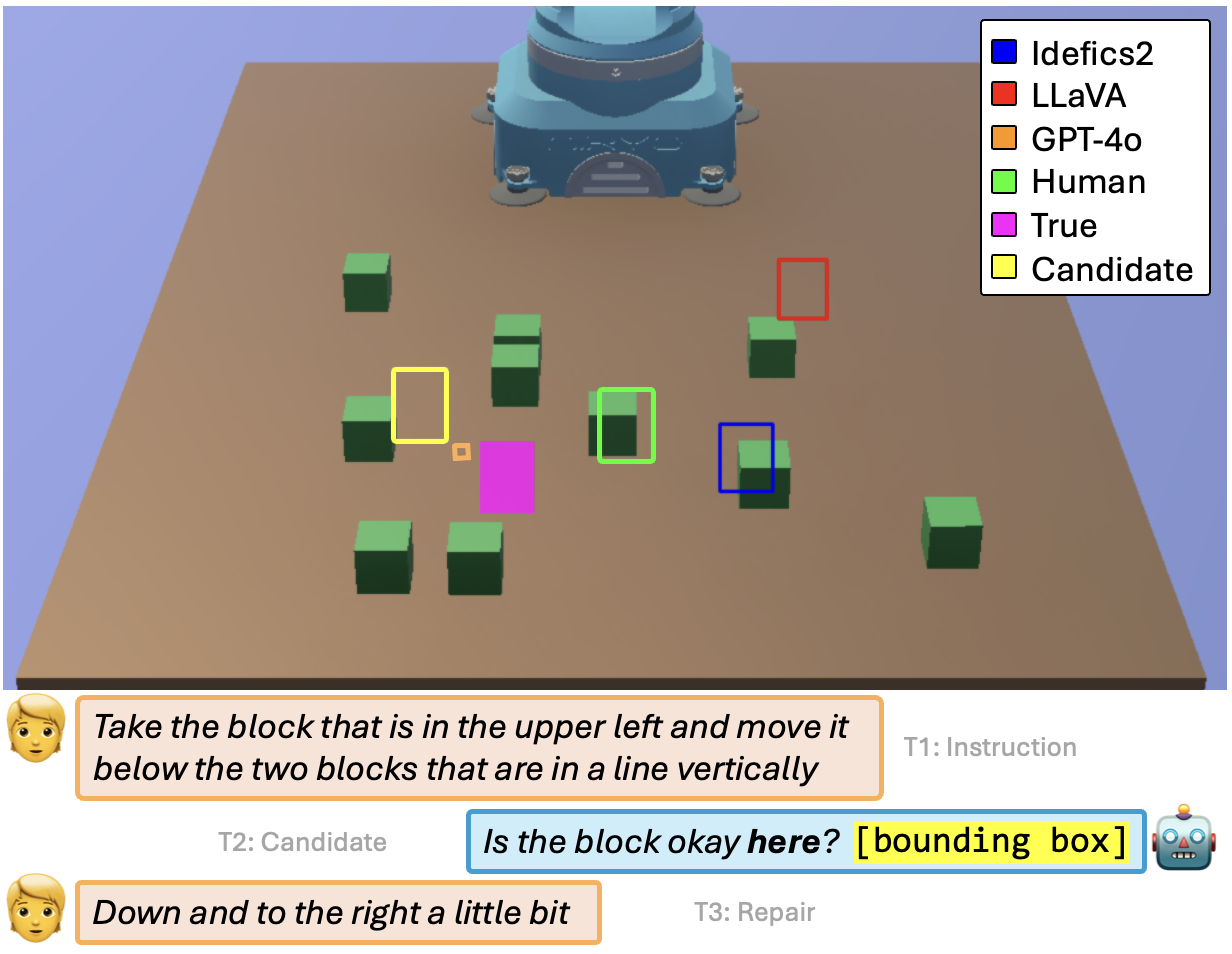}
    \end{subfigure}
    
    \caption{Two medium-difficulty dialogues with the bounding boxes predicted by the VLMs and humans.}
    \label{fig:model_comparison}
\end{figure}

We further analyse 50 \DatasetShort{} dialogues to understand the proficiency differences of VLMs with that of human participants (see \Cref{fig:model_comparison} for some examples). We find that humans and GPT-4o tend to have similar predictions when processing candidate responses and their repairs. Models are usually able to process simple spatial repairs (i.e., left, right, above) \cite{chiyah-garcia-etal-2023-referring}, however, they particularly struggle with more abstract concepts (i.e., rows, columns) \cite{ilinykh-etal-2022-look}. 
In that regard, both Idefics2 and LLaVA exhibit comprehension of the simpler referring expressions, but have a challenging time with longer or more complex sequences. Idefics2 also seems prone to over-correct (e.g., \textit{right} pushes predictions to the absolute right). We also notice that VLMs sometimes predict bounding boxes that do not align with blocks in the source prediction task, which is unexpected given their pre-training (and fine-tuning) with referring expression objectives.
For the target prediction task, VLMs significantly struggle and do not process repairs. GPT-4o is the only exception, which shows a more fine-grained understanding of the task yet falls behind on more difficult dialogues.

\section{Conclusion}

This paper explores VLM's capabilities of handling repairs with a new dataset, \DatasetLong. We collect 795 collaborative dialogues in a highly ambiguous tabletop manipulation environment that requires complex multi-modal task instructions as well as repairs. In these scenarios, referential ambiguities are common and thus being able to process TPRs becomes essential to complete the task. We validate these dialogues through an in-person human study where participants attempt to solve the task, providing a baseline for model performance. 

We then show that VLMs struggle to process TPRs alongside other dialogues out of the box and propose improving their fine-tuning regime with different loss criteria. Our results indicate that VLMs benefit from partially masking the input tokens when learning to process TPRs along with the task, resulting in more generalisable models across single-turn instructions and TPRs for one of the tasks. However, this needs to be balanced as masking too many tokens hurts models when data is smaller due to a lack of training signal.

We finish with an error analysis comparing these models and human participants, finding stark differences in how they process repairs. Surprisingly, models struggle with the dialogues that human participants found easiest, highlighting a large gap in their ability to process TPRs, despite these being fundamental to effective dialogue coordination.
We believe that future research should focus on two major improvements for VLMs: 1) design training objectives that facilitate \emph{learning from interaction}, including TPRs; 2) equip models with more object-centric visual representations that facilitate visual grounding tasks \cite{parekh2024investigatingroleinstructionvariety} which are of paramount importance for situated collaborative tasks. 






\section{Limitations}

One of the limitations of the paper is that we collected the data for the dataset as an annotation task. This allowed us to divide the task into instances that participants could annotate separately (instructions or corrections), which makes the annotation conceptually easier to carry out and mimics how the initial Block World instructions were collected. This allowed us to focus on the dialogue phenomena that we care about, corrections in situated contexts, at the cost of the overall collaborative nature of the task. Ideally, it would be better to design a Wizard-of-Oz or human-human experiment to collect the annotations during the interaction, at a much greater time and cost. 

We also acknowledge that our environment is visually simple compared to images that can be found on the Internet which are commonly used for pretraining VLMs. However, we don't consider this as a weakness but as a strength: our \DatasetLong{} is built using the Unity game engine and allows us to test models' ability to process TPRs ignoring other confounding variables such as processing complex visual scenes.

Another limitation of our work is that we do not attempt to model the entire spectrum of a robotic manipulation task \cite[e.g., ][]{octo_2023}. Instead, we follow previous work in Embodied AI (e.g., ALFRED~\cite{shridhar2020alfred}, Simbot Arena~\cite{gao2023alexaarenausercentricinteractive}, \emph{inter alia}) which casts object manipulation as API call generation of an action label combined with a bounding box around the object to be manipulated. Although we don't generate actions, the prediction tasks that we explore in \DatasetShort{} involve generating the coordinates of a bounding box that can be used for picking up a specific block or putting it down onto a specific box on the board.

To the best of our knowledge, we have used state-of-the-art VLMs that are open-weight and can be easily used for fine-tuning and inference. However, due to the fast pace of research in this space, it is possible that by the time reviewers read this manuscript, a new open-weight VLM will be announced. For this reason, we decided to also report the performance of a frontier model such as GPT-4o which has superior performance to many text-only as well as visual+language tasks.

\section{Potential Risks}

Embodied AI and human-robot collaboration offer exciting possibilities, but also introduce a new range of potential risks. Embodied AI systems must be designed and implemented with rigorous safety protocols to minimise the risk of accidents for human workers. This includes robust sensor systems, clear communication protocols, and fail-safe mechanisms to prevent harm in case of malfunctions. In this paper, we do not use an actual robot but we simulate the use case of a robot arm operating in a factory and collaborating with a user to complete a pick\&place task. In this scenario, we argue that it is essential that the robot is able to process clarification exchanges, and TPRs are an important mechanism to ensure that models can robustly correct mistakes when they arise.

\section*{Acknowledgements}

Chiyah-Garcia’s PhD is funded under the EPSRC iCase with Siemens (EP/T517471/1). We thank the anonymous reviewers for their insightful comments that helped improve this paper. Finally, the authors acknowledge the
use of the Heriot-Watt University high-performance computing facility (DMOG) and associated support services in the completion of this work.

\bibliography{bib/anthology,bib/all,bib/custom}

\clearpage

\appendix

\section{Related Datasets}\label{ap:related_work}

We compare \DatasetLong{} with related works in \Cref{tab:dataset_comparison}. Our work builds on a robotic manipulation task involving common pick\&place actions and rich referring expressions in a situated dialogue. Similar to our benchmark, both TEACh \cite{Padmakumar2022} and DialFRED \cite{gao2022dialfred} extend the popular ALFRED \cite{shridhar2020alfred} to include dialogues with clarificational exchanges. From a high-level ALFRED directive (e.g., ``prepare coffee''), these works crowd-source possible questions for low-level actions that may be underspecified in the original task (e.g., ``where is the coffee mug?''). Thus, they often operate on the last level of \citeposs{Clark96} joint action ladder, \textit{action and consideration}, where the goal is already clear and lack of knowledge (underspecified location of objects or actions available) represents the primary ambiguity \cite[e.g., VIMA-Bench; ][]{pmlr-v202-jiang23b}). These works rarely feature bi-directional information flow ($human \leftrightarrows agent$) in their dialogues, unlike \DatasetShort{} which ensures dialogue dependency with partial robot movements and candidate responses. 
DialFRED, Alexa Arena \cite{gao2023alexaarenausercentricinteractive} and CoDraw iCRs \cite{madureira-schlangen-2023-instruction} contain questions for underspecified actions/objects (e.g., ``Which spoon should I pick up?'') but, in these situations, the ambiguity is reduced to a minimum so that candidates are easily distinguishable with strong visual representations (e.g., ``yellow or blue spoon?'' or ``left or right?''). 

By contrast, SIMMC 2.0 \cite{kottur-etal-2021-simmc} includes a large number of visually ambiguous objects (e.g., 5 identical red t-shirts in view) within long dialogues that reference multiple items (averaging 4.5±2.4 unique objects per dialogue). This makes it rich in cross-modal coordination phenomena, with terse referring expressions that mix spatial, historical and visual cues (e.g., ``Pick the red shirt. Which one? The one on the right wardrobe, above the blue jumper''). Similarly, Cups \cite{Dobnik2020} aligns with our work, as it also explores coordination in highly ambiguous environments where many objects share the same shape/colour but are in different positions, requiring complex referring expressions. However, Cups focuses on Frame of Reference coordination rather than manipulation or resolving coreferences. \DatasetShort{} is aimed at exploring this coordination as part of a collaborative human-robot tabletop manipulation task.




\begin{table*}[htb]
\centering
\addtolength{\tabcolsep}{-0.15em}
\renewcommand{\arraystretch}{1.3}

\small{

\begin{tabular}{@{}l|c|c|c|c|l@{}}
\toprule

\textbf{Related Work}                                      & \textbf{Manipulation} & \textbf{Ref Exp}  & \textbf{Dialogue}   & \textbf{Repairs}    & \textbf{Miscommunication}  \\

\midrule
VIMA-Bench \cite{pmlr-v202-jiang23b}                       & \greentick            & \relationaltype           & \redcross           & \redcross           & -                  \\
\rowcolor[HTML]{EFEFEF} ALFRED \cite{shridhar2020alfred}                                     & \greentick            & \relationaltype           & \redcross           & \redcross           & -                  \\
TEACh \cite{Padmakumar2022}                                & \greentick            & \visualtype\relationaltype          & \greentick          & \greentick          & (4) Action                 \\
\rowcolor[HTML]{EFEFEF} DialFRED \cite{gao2022dialfred}                            & \greentick            & \visualtype\relationaltype          & \greentick          & \greentick          & (4) Action                 \\
Alexa Arena \cite{gao2023alexaarenausercentricinteractive} & \greentick            & \visualtype\relationaltype          & \greentick          & \redcross           & (4) Action                 \\
\rowcolor[HTML]{EFEFEF} SIMMC 2.0 \cite{kottur-etal-2021-simmc}                    & \redcross             & \visualtype\relationaltype\dialoguetype          & \greentick          & \greentick          & (3) Understanding          \\

CoDraw iCR \cite{madureira-schlangen-2023-instruction}	&    \redcross & 	\visualtype\relationaltype & 	\greentick &	\greentick  &	(4) Action \\

\rowcolor[HTML]{EFEFEF} Cups \cite{Dobnik2020}                                     & \redcross             & \visualtype\relationaltype\dialoguetype        & \greentick          & \greentick          & (3) Understanding          \\

Block World \cite{bisk-etal-2016-natural}                  & \greentick            & \visualtype\relationaltype          & \redcross           & \redcross           & -                  \\
\midrule
\textbf{\DatasetLong~(Ours)}                          & \textbf{\greentick}   & \visualtype\relationaltype & \textbf{\greentick} & \textbf{\greentick} & \textbf{(3) Understanding} \\ 
\bottomrule
\end{tabular}
}
\caption{Comparison of related works to our dataset \DatasetShort{}. Columns describe whether the datasets have: (1 \textbf{Manipulation}) robotic manipulation as a task (i.e., pick\&place); (2 \textbf{Ref Exp}) referring expression type following \cite{chiyah-garcia-etal-2023-referring}, \relationaltype for relational expressions (e.g., ``the left one''), \visualtype for references to visual properties (e.g., ``the green one'') or \dialoguetype when referring to the dialogue history (e.g., ``the one I mentioned''); (3 \textbf{Dialogue}) dialogue with multiple turns for the same action; (4 \textbf{Repairs}) explicit repairs in the data (i.e., clarifications or corrections); and (5 \textbf{Miscommunication}) the miscommunication level based on \citeauthor{Clark96}'s (\citeyear{Clark96}) joint action ladder.} 
\label{tab:dataset_comparison}
\end{table*}

\section{\DatasetLong}\label{ap:dataset}

This section provides additional details about the dataset.

\subsection{AMT Collection}\label{ap:dataset_collection}

\begin{figure}[ht]
  \centering
  \includegraphics[width=\linewidth]{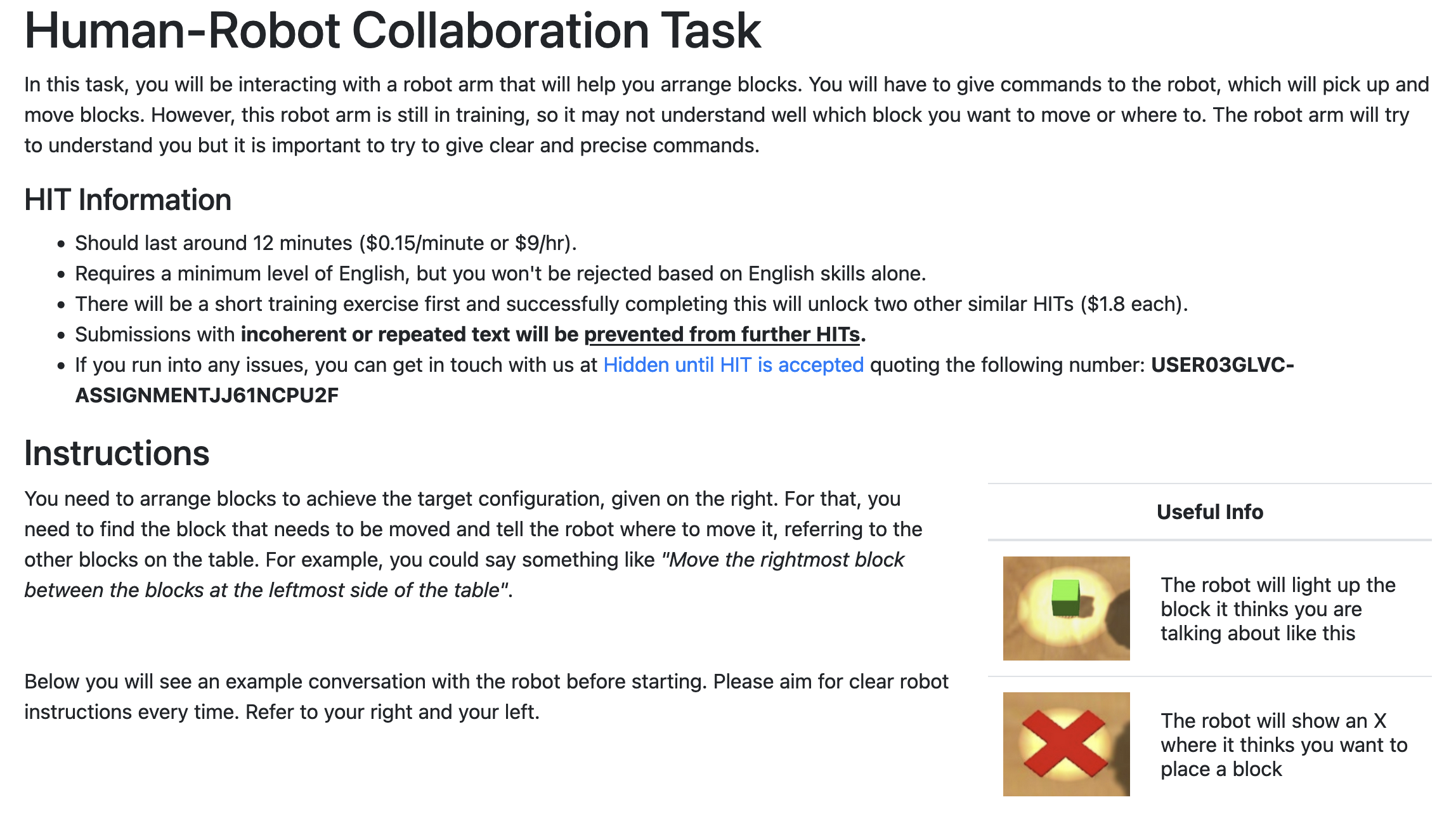}
  \caption{Instructions that AMT workers saw in the landing page of the HIT.}
  \label{fig:instructions}
\end{figure}
\begin{figure}[ht]
  \centering
  \includegraphics[width=\linewidth]{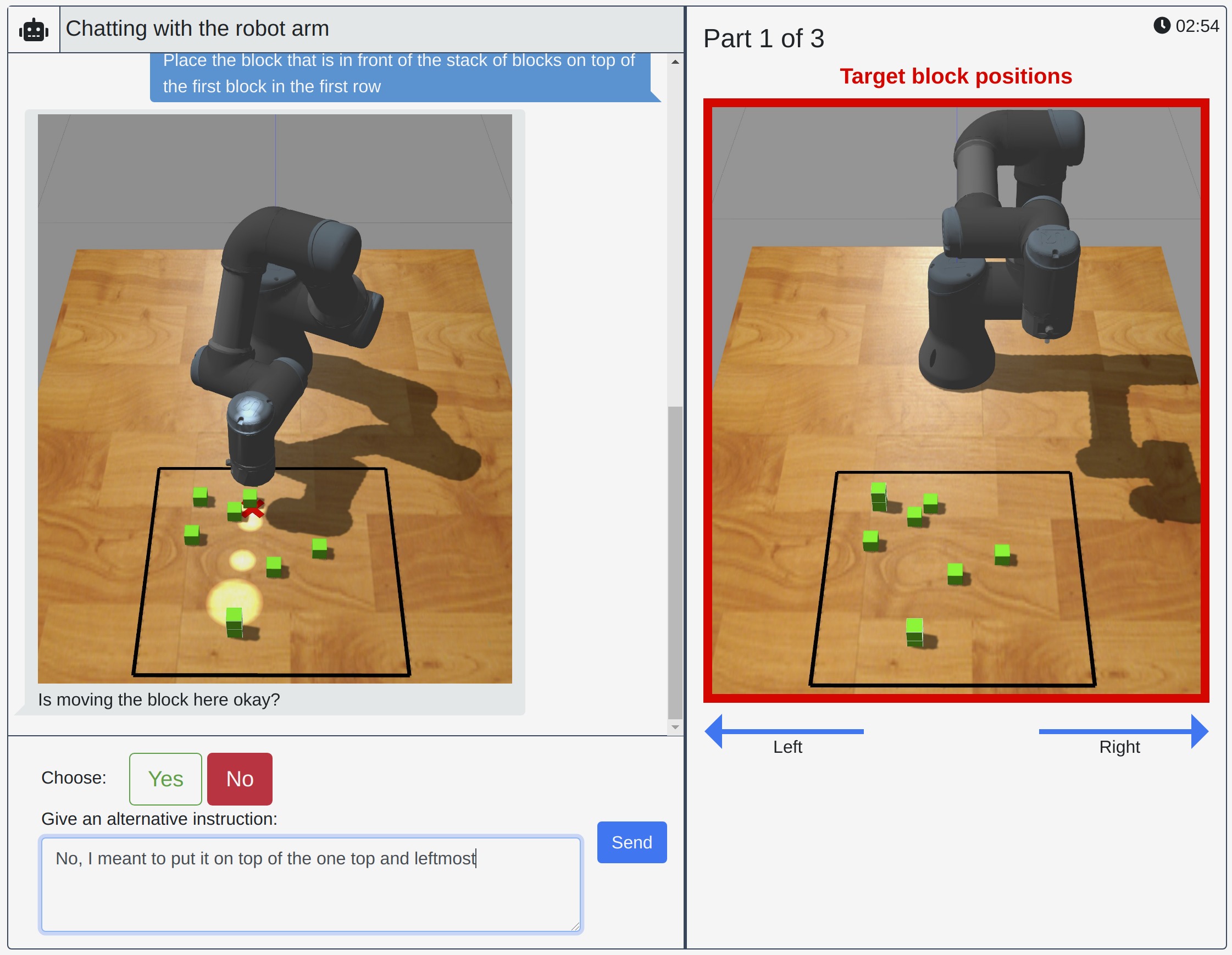}
  \caption{User interface for the data collection. The images of the robot and blocks would change depending on the task or the instructions given. The left has the chat and a text box to send messages whereas the right shows how the final placement should look in the end.}
  \label{fig:data_collection_interface}
\end{figure}

Each task or Human Intelligence Task (HIT) in AMT consisted of 4 dialogues, with an initial easier, training dialogue that workers were free to experiment with. AMT workers could repeat the task for as long as there were block configurations they had not seen before. The instructions are in \Cref{fig:instructions}, with the user interface of the task itself in \Cref{fig:data_collection_interface}.
We introduced many attention checks, particularly with the repairs, where workers were asked to select whether the robot had performed the correct action (Yes/No) before entering free-text repairs. If their answer did not match the expected answer too many times, they would get a warning and be limited from submitting any subsequent HITs. We used the CRWIZ framework \cite{chiyah-garcia-etal-2020-crwiz} to control many of these details, including the robot's non-verbal actions (i.e., movements, pointing), effectively running a simulation or sequence of events different to each worker in AMT.

At the end of the task, workers gave free-form comments and feedback in scales about how easy the task was and the Godspeed Questionnaire IV for Perceived Intelligence \cite{Bartneck2009godspeed} to ensure that the task was not too difficult and that the agent was not too unintelligent (due to the continuous ``misunderstandings'').

We limited HITs to 30 minutes and the mean elapsed time for all the HITs was 12 minutes (SD=4.5 minutes), or 4 minutes per dialogue. The payment was \$1.8 per HIT, so workers were paid, on average, a rate of \$9.0/hr. This pay is above the Federal minimum wage in the US (\$7.25/hr or \$0.12/min) at the time of the data collection from May to September 2021. 178 unique AMT workers submitted HITs for our task (58 female, 106 male, and 12 rest) and we limited our task to be only available in the United States. We manually checked data subsets for offensive, toxic language, or personal information. Participants signed a virtual consent with details about the data collection, how the data would be used and how they could withdraw if they wished to do so. We did not collect any personal or sensitive information. Ethics approval for this collection was provided by our institution's ethics committee.


\subsection{Human Baselines}

\begin{figure}[htb]
  \centering
  \includegraphics[width=\linewidth]{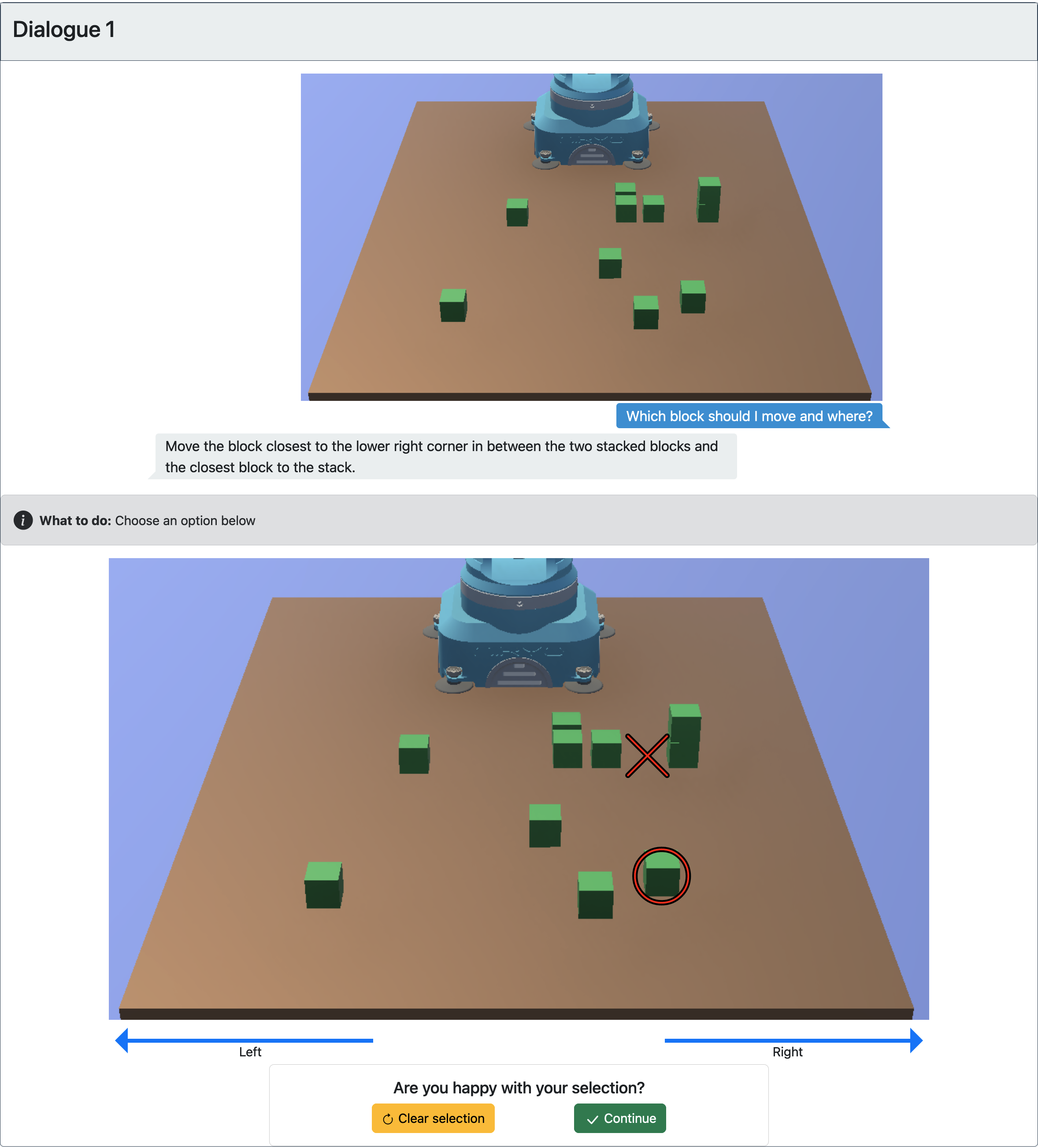}
  \caption{User interface for the human validation. Participants had to click on the large image at the bottom where they thought that the instructions referred to, both for the source block and the target position.}
  \label{fig:human_study}
\end{figure}

Participants interacted with the interface shown in \Cref{fig:human_study}, where the chat was already filled by a randomly selected dialogue. Participants had to read the instructions and repairs and provide their best interpretation of which block the instructions referred to and where to place it. Once satisfied with their choice, they would confirm and proceed to the next dialogue until the session ended. This interface uses the Unity engine (v2022.3)\footnote{\url{https://unity.com/}} and \cite{de2024planning} to generate environments based on the original Block World configurations.

The study ran from January to April 2024 and it lasted around 40 minutes. We recruited a total of 22 participants (5 female, 17 male, most common age range was 23-29 and above undergraduate education level) through internal institution mailing lists. 12 were native English speakers and most participants self-reported a near bilingual level of English proficiency. On average, participants annotated 50 dialogues each and received £10 for taking part in the study (a rate of £15/hour). This pay is above the minimum wage in the UK at the time of the study (£11.44/hour). Participants signed consent forms with details about how the data would be used and published as well as the procedure to withdraw if they wished so. We did not collect any personal or sensitive information. 
Ethics approval for this study was provided by our institution's ethics committee.

We collected 991 after discarding 61 low-quality annotations. The annotations are split as follows: 343 for instructions (34.6\%), 327 for source block TPRs (33.0\%) and 321 for target position TPRs (32.4\%).

\subsection{Dataset Statistics}

\Cref{tab:collected_entries_summary} provides additional statistics for the \DatasetLong{} dataset.

\begin{table}[ht]
    \centering
    \begin{tabular}{l l}
        \toprule
        \textbf{Attribute} & \textbf{Value} \\
        \midrule
        Instructions & 795 (38.6\%) \\
        Source TPRs & 629 (30.5\%) \\
        Target TPRs & 635 (30.8\%) \\
        \midrule
        Total Entries & 2059 \\
        \bottomrule
    \end{tabular}
    \caption{Summary of collected entries in \DatasetShort.}
    \label{tab:collected_entries_summary}
\end{table}

For our experiments, we join the instructions from the original Block World with the collected entries to obtain a larger dataset, see \Cref{tab:dataset_split_summary}. The initial instructions in our dialogues are the same as in \cite{bisk-etal-2016-natural}, as they contain both which block to move and where to move it. We also use the same block arrangements as in the original dataset, respecting train/test splits. Block configurations are unique across train-test splits, so models are tested on unseen world arrangements.   

\begin{table*}[ht]
    \centering
    \begin{tabular}{lcccc}
        \toprule
        \textbf{Split} & \textbf{Train} & \textbf{Dev} & \textbf{Test} & \textbf{Total} \\
        \midrule
        Block World \cite{bisk-etal-2016-natural} & 2403 (66.5\%) & 360 (100.0\%) & 810 (48.8\%) & 3573 \\
        \DatasetLong{} & 1210 (33.5\%) & 0 (0.0\%) & 849 (51.2\%) & 2059 \\
        \midrule
        Experiment Data & 3613 & 360 & 1659 & 5632 \\
        \bottomrule
    \end{tabular}
    \caption{Data used in this paper. We use the train/test sets during our experiments of Sections 4, 5 and 6.}
    \label{tab:dataset_split_summary}
\end{table*}



\section{Experimental Setup}\label{ap:experimental_setup}

\subsection{Prompts}

Each task has its own slightly different prompt, see \Cref{tab:task_prompts}. We also provide example full prompts used during zero-shot and fine-tuning in \Cref{tab:example_prompts}.

When parsing model generations, we try to parse a bounding box with 4 decimal numbers. However, models sometimes generate invalid boxes or additional tokens, particularly in zero-shot settings. In these cases, less than 3\%, we count this prediction as `failed' and use the output of the random baseline to calculate other metrics. Idefics2 and GPT-4o are more prone to generate additional tokens.

\begin{table}[ht]
    \centering
    \begin{tabular}{m{1.5cm} m{5cm}}
        \toprule
        \textbf{Task} & \textbf{Task Instruction} \\
        \midrule
        Source Block Selection & Answer only with the bounding box of the block mentioned. The bounding box consists of 4 values between 0 and 1. Here is an example: [0.000, 0.123, 0.075, 0.204] \\
        Target\newline Position\newline Prediction & Answer only with the final bounding box location mentioned. The bounding box consists of 4 values between 0 and 1. Here is an example: [0.000, 0.123, 0.075, 0.204] \\
        \bottomrule
    \end{tabular}
    \caption{Task instructions used to prompt models.}
    \label{tab:task_prompts}
\end{table}

\begin{table*}[ht]
    \centering
    \begin{tabularx}{\textwidth}{lX}
    \toprule
        \textbf{Model} & \textbf{Sample} \\
        \midrule
        Idefics2 8B & User: <image> Answer only with the bounding box of the block mentioned. The bounding box consists of 4 values between 0 and 1. Here is an example: [0.000, 0.123, 0.075, 0.204]. take the block in the very middle and move it slightl left and down.<end\_of\_utterance> \newline Assistant: location bounding box [0.329, 0.724, 0.339, 0.741]
I recall you wanting the block here, is this correct?<end\_of\_utterance> \newline User: move it to the right a little<end\_of\_utterance> \newline Assistant: \\
        & \cellcolor[HTML]{EFEFEF} \texttt{Model generation:} location bounding box [0.394, 0.585, 0.436, 0.682] \\
        \\
        LLaVA v1.5 7B & <s> A chat between a curious user and an artificial intelligence assistant. The assistant gives helpful, detailed, and polite answers to the user's questions. Answer only with the bounding box of the block mentioned. The bounding box consists of 4 values between 0 and 1. Here is an example: [0.000, 0.123, 0.075, 0.204]. \newline USER: <image> There is a block in the center of the board. Take that block and move it to the left in front of the other block. \newline ASSISTANT: block bounding box block bounding box [0.391, 0.635, 0.434, 0.735] Can you confirm it is this block?</s> \newline USER: The block that is a row closer to you and to the right. \newline ASSISTANT:\\
        & \cellcolor[HTML]{EFEFEF} \texttt{Model generation:} block bounding box [0.391, 0.635, 0.434, 0.735] \\
        \\
        GPT-4o & System: You are a helpful assistant that responds with only with the bounding box of the location mentioned. The bounding box consists of 4 values between 0 and 1. Here is an example: [0.000, 0.123, 0.075, 0.204]. \newline User: <image> Take the block that is furthest to the right in the fourth row and place it in front of  all the blocks on the table. \newline Assistant: block bounding box [0.749, 0.714, 0.800, 0.818] Is it this block? \newline User: Take the block that is directly above this block and to the left of it. \\
        & \cellcolor[HTML]{EFEFEF} \texttt{Model generation:} block bounding box [0.624, 0.571, 0.675, 0.675] \\
        
        \bottomrule
    \end{tabularx}
    \caption{Full prompts given to models, including special tokens.}
    \label{tab:example_prompts}
\end{table*}




\subsection{Model Fine-Tuning}

For our experiments, we use the default hyperparameters for each model (Idefics2 and LLaVA). We train for 1 epoch as we did not observe benefits from training for longer. Images had the size of 1024x576 pixels, and we used the model's default image processing pipelines. 
Training batches had both instruction and repair examples, sampled at random. We did not use custom sampling to balance the type of entries in batches and leave this for future work. 

We fine-tune the VLMs with the recommended parameter-efficient methods: LoRA~\cite{hu2021lora} for LLaVA and QLoRA \cite{NEURIPS2023_1feb8787} for Idefics2. 
We used 2x NVIDIA A40 (40GB) at most and set the maximum sequence length to 2048.
Each experiment (training and testing) takes around 1 hour to fully complete. The final table of experiments takes approximately 30 GPU hours to run.

\section{Additional Error Analysis}\label{ap:error_analysis}

As mentioned in \Cref{sec:error}, it seems that easier referring expressions or TPRs have a lower number of words on average than harder ones (see \Cref{tab:words_per_dialogue}). 

We also provide in \Cref{tab:difficulty_levels_gpt4o} the results of selecting GPT-4o performance as the proxy for example difficulty. Since we only have 1 prediction value for each \DatasetShort{} entry and we cannot do the average as with the human annotations, we instead order the data by distance (source and target separately) and divide it into three equal sizes at the 33rd and 66th percentiles: 1) \textbf{Easy} category contains the top 33\% of entries (shortest distance to source/target); 2) \textbf{Medium} contains entries that are between the 33\% and 66\% distances; and 3) \textbf{Hard} is reserved for the worst performing 33\% entries.  

We find that Idefics2 and LLaVA have low performance on all levels, whilst GPT-4o only surpasses humans on easy dialogues (not surprising), but quickly deteriorates to distances worse than random on hard dialogues. Interestingly, humans perform well across all difficulties except the hard target predictions. We also observe that the mean word trends get stronger (\Cref{tab:words_per_dialogue}), although we leave the analysis to prove a correlation for future work. These results further show that models do not process TPRs as consistently as humans.

\begin{table}[ht]
    \centering
    \small{
    \begin{tabular}{lcc}
        \toprule
        \textbf{Difficulty Level} & \textbf{Source Block} & \textbf{Target Position} \\
        \midrule
        \multicolumn{2}{l}{\textit{Human Performance}} \\
        ~Easy & 28.69 (±12.20) & 32.19 (±14.66) \\
        ~Medium & 28.80 (±12.97) & 31.27 (±12.86) \\
        ~Hard & 27.62 (±12.19) & 27.46 (±12.35) \\

        \midrule
        \multicolumn{2}{l}{\textit{GPT-4o Performance}} \\

        ~Easy   & 29.95 (±12.09) & 34.29 (±13.81) \\
        ~Medium & 29.62 (±12.02) & 31.46 (±12.45) \\
        ~Hard   & 26.52 (±12.31) & 24.54 (±11.73) \\

        \bottomrule
    \end{tabular}}
    \caption{Mean and Standard Deviation (SD) for test dialogues in \DatasetShort{} by difficulty level and task. We differentiate between using human performance (\Cref{sec:error}) and GPT-4o performance (\Cref{ap:error_analysis}) as the difficulty proxies.}
    \label{tab:words_per_dialogue}
\end{table}

\begin{table}[t]
    \centering
    \addtolength{\tabcolsep}{-0.3em}
    \footnotesize{
    \begin{tabular}{@{}ll|cc|c@{}}
    \toprule
    \multirow{2}{*}{\textbf{Difficulty}} & \multirow{2}{*}{\textbf{Model}}         & \multicolumn{2}{c}{\textbf{Source}}     & \textbf{Target} \\
    \textbf{}          &      & \textbf{Acc ↑} & \textbf{Mean (SD) ↓} & \textbf{Mean (SD) ↓} \\
    \midrule
    
    Easy & Idefics2            & 0.43 & 2.60 (±2.6) & 6.11 (±2.9) \\
    & LLaVA          & 0.48 & 2.02 (±2.0) & 4.20 (±2.9) \\
    & Humans   & 0.78 & 1.07 (±1.1) & 2.05 (±2.1) \\
    & GPT-4o                 & 0.93 & .00 (±.0)   & 1.27 (±.5)  \\
    
    \midrule
    Medium                   & Idefics2            & 0.47 & 2.29 (±2.3) & 4.97 (±2.6) \\
    & LLaVA          & 0.47 & 2.01 (±2.0) & 4.25 (±2.7) \\
    & Humans & 0.77 & 1.05 (±1.1) & 2.23 (±2.3) \\
    & GPT-4o                 & 0.58 & 1.26 (±1.3) & 2.63 (±.4)  \\
    \midrule 
    Hard                     &  Idefics2            & 0.04 & 4.91 (±4.9) & 5.92 (±3.1) \\
    &LLaVA          & 0.22 & 4.61 (±4.6) & 5.69 (±3.4) \\
    & Humans   & 0.64 & 2.52 (±2.5) & 4.84 (±3.7) \\
    & GPT-4o                 & 0.00 & 8.33 (±8.3) & 6.31 (±3.1) \\
    
    \bottomrule
    \end{tabular}}
    \caption{Model performances across subsets with increasing levels of complexity (\textit{easy, medium, hard}) using GPT-4o performance as the difficulty proxy.}
    \label{tab:difficulty_levels_gpt4o}
\end{table}

\section{Additional Experimental Results}
\label{ap:additional_results}

We completed a comprehensive set of ablations testing for different ways of masking (as defined in the main paper), different VLMs, and different data regimes as well. Table \ref{tab:additional_results} presents the full set of results.

\begin{table*}[ht]
\centering
\addtolength{\tabcolsep}{-0.2em}
\small{
\begin{tabular}{@{}clll|cc|c@{}}
\toprule
\textbf{Test}           & \textbf{Train} & \multirow{2}{*}{\textbf{Model}}           & \multirow{2}{*}{\textbf{Loss}} & \multicolumn{2}{c|}{\textbf{Source}}              & \textbf{Target}          \\
\textbf{Data}                    &  \textbf{Data}              &                          &               & \textbf{Accuracy ↑} & \textbf{Mean Distance (SD) ↓} & \textbf{Mean Distance (SD) ↓} \\
\midrule
\multicolumn{1}{l}{}                    &                & Idefics2 8B \textit{zeroshot}   & Default           & 0.18                      & 5.06 (±3.50)             & 9.55 (±2.50)             \\
\multirow{12}{*}{\rotatebox[origin=c]{90}{\textbf{Instructions}}} & Instructions   & Idefics2 8B   & Default          & 0.33                      & 3.33 (±3.02)             & 4.93 (±2.64)             \\
                                        & Instructions   & Idefics2 8B   & User-turn     & 0.33                      & 3.33 (±3.02)             & 4.93 (±2.64)             \\
                                        & Instructions   & Idefics2 8B   & Completions           & 0.32                      & 3.44 (±3.04)             & 4.75 (±2.62)             \\
                                        & Repairs    & Idefics2 8B   & Default          & 0.21                      & 4.36 (±3.13)             & 5.38 (±2.44)             \\
                                        & Repairs    & Idefics2 8B   & User-turn     & 0.22                      & 4.59 (±3.25)             & 8.03 (±3.21)             \\
                                        & Repairs    & Idefics2 8B   & Completions           & 0.14                      & 4.88 (±3.07)             & 6.19 (±2.97)             \\
                                        & Full           & Idefics2 8B   & Default          & 0.30                      & 3.39 (±2.92)             & 4.75 (±2.60)             \\
                                        & Full           & Idefics2 8B   & User-turn     & 0.36                      & 3.12 (±3.00)             & 4.59 (±2.59)             \\
                                        & Full           & Idefics2 8B   & Completions           & 0.33                      & 3.33 (±2.99)             & 4.52 (±2.58)             \\
                                        &                & LLaVA v1.5 7B \textit{zeroshot} & Completions           & 0.24                      & 4.40 (±3.34)             & 5.74 (±3.08)             \\
                                        & Instructions   & LLaVA v1.5 7B & Default          & 0.25                      & 3.77 (±2.99)             & 5.06 (±2.92)             \\
                                        & Instructions   & LLaVA v1.5 7B & User-turn     & 0.27                      & 3.79 (±3.03)             & 5.02 (±2.86)             \\
                                        & Instructions   & LLaVA v1.5 7B & Completions           & 0.27                      & 3.76 (±3.00)             & 4.40 (±2.69)             \\
\multicolumn{1}{l}{}                    & Repairs    & LLaVA v1.5 7B & Default          & 0.14                      & 4.97 (±3.24)             & 6.24 (±2.91)             \\
\multicolumn{1}{l}{}                    & Repairs    & LLaVA v1.5 7B & User-turn     & 0.18                      & 4.19 (±2.82)             & 6.37 (±2.91)             \\
\multicolumn{1}{l}{}                    & Repairs    & LLaVA v1.5 7B & Completions           & 0.09                      & 5.12 (±3.00)             & 6.27 (±2.86)             \\
\multicolumn{1}{l}{}                    & Full           & LLaVA v1.5 7B & Default          & 0.22                      & 3.82 (±2.91)             & 4.46 (±2.49)             \\
\multicolumn{1}{l}{}                    & Full           & LLaVA v1.5 7B & User-turn     & 0.24                      & 3.65 (±2.86)             & 4.60 (±2.54)             \\
\multicolumn{1}{l}{}                    & Full           & LLaVA v1.5 7B & Completions           & 0.19                      & 3.93 (±2.73)             & 4.51 (±2.82)             \\
\multicolumn{1}{l}{}                    &                & GPT-4o \textit{zeroshot}        &           & 0.30                      & 3.83 (±3.38)             & 4.22 (±2.65)             \\
\midrule

\multicolumn{1}{l}{}                    &                & Idefics2 8B \textit{zeroshot}   & Completions           & 0.00                      & 3.74 (±1.95)             & 4.13 (±2.05)             \\
\multirow{18}{*}{\textbf{\rotatebox[origin=c]{90}{Repairs}}}  & Instructions   & Idefics2 8B   & Default          & 0.21                      & 4.19 (±2.75)             & 5.11 (±2.60)             \\
                                        & Instructions   & Idefics2 8B   & User-turn     & 0.26                      & 3.74 (±2.84)             & 4.68 (±2.12)             \\
                                        & Instructions   & Idefics2 8B   & Completions           & 0.28                      & 3.66 (±2.77)             & 3.98 (±2.58)             \\
                                        & Repairs    & Idefics2 8B   & Default          & 0.58                      & 2.19 (±3.06)             & 7.07 (±2.77)             \\
                                        & Repairs    & Idefics2 8B   & User-turn     & 0.58                      & 2.16 (±2.92)             & 6.45 (±2.22)             \\
                                        & Repairs    & Idefics2 8B   & Completions           & 0.57                      & 1.55 (±1.95)             & 7.35 (±2.02)             \\
                                        & Full           & Idefics2 8B   & Default          & 0.26                      & 3.43 (±2.40)             & 5.81 (±2.35)             \\
                                        & Full           & Idefics2 8B   & User-turn     & 0.32                      & 2.94 (±2.56)             & 6.00 (±1.95)             \\
                                        & Full           & Idefics2 8B   & Completions           & 0.47                      & 2.29 (±2.36)             & 5.90 (±2.71)             \\
                                        &                & LLaVA v1.5 7B \textit{zeroshot} & Completions           & 0.13                      & 3.29 (±2.09)             & 3.45 (±1.20)             \\
                                        & Instructions   & LLaVA v1.5 7B & Default          & 0.18                      & 3.43 (±2.49)             & 3.49 (±1.34)             \\
                                        & Instructions   & LLaVA v1.5 7B & User-turn     & 0.16                      & 3.47 (±2.42)             & 3.46 (±1.39)             \\
                                        & Instructions   & LLaVA v1.5 7B & Completions           & 0.25                      & 2.79 (±2.47)             & 3.76 (±1.88)             \\
                                        & Repairs    & LLaVA v1.5 7B & Default          & 0.07                      & 3.20 (±1.79)             & 3.67 (±1.31)             \\
                                        & Repairs    & LLaVA v1.5 7B & User-turn     & 0.13                      & 3.28 (±2.05)             & 3.89 (±1.35)             \\
                                        & Repairs    & LLaVA v1.5 7B & Completions           & 0.50                      & 2.33 (±2.64)             & 5.57 (±1.67)             \\
                                        & Full           & LLaVA v1.5 7B & Default          & 0.44                      & 2.66 (±2.76)             & 3.69 (±2.18)             \\
                                        & Full           & LLaVA v1.5 7B & User-turn     & 0.37                      & 2.69 (±2.48)             & 4.04 (±2.41)             \\
\multicolumn{1}{l}{}                    & Full           & LLaVA v1.5 7B & Completions           & 0.54                      & 2.01 (±2.47)             & 4.56 (±2.69)             \\
\multicolumn{1}{l}{}                    &                & GPT-4o \textit{zeroshot}        & Default          & 0.41                      & 2.75 (±3.06)             & 2.95 (±2.17)       \\
\bottomrule
\end{tabular}}

\caption{Model performance in source and target sub-tasks across all losses.}
\label{tab:additional_results}
\end{table*}

\section{Other Notes}

\paragraph{Object detection and predictions}
Splitting the task into object detection and then prediction out of a list of candidates did not show any advantages in our initial tests, possibly because these models are already pretrained with object detection objectives \cite{liu2023improvedllava,laurençon2024matters}. We already see better performance with source block prediction than with target location prediction. Thus, we unified both tasks into a single common task format (bounding box prediction) for easier comparison. Additionally, we would like to point out that target prediction would be a harder task because it requires the model to derive a bounding box for an empty board location, which models particularly struggle as shown in \Cref{sec:error}.

\paragraph{Repairs facilitate the task}
Repairs are fundamental capabilities in human conversations, which our human study further shows with the increasing performance with TPRs (see \Cref{tab:human_performance}). 
TPRs have the dual role of narrowing down the candidate pool (ie., ``not this block'') and providing new information to uniquely identify the correct referent (ie., ``just left of there''). Our paper shows that current out-of-the-box models do not have the capabilities to exploit repairs to facilitate the task (unlike humans), but we should try to distil these skills with alternative training regimes that facilitate learning from interaction data. Handling TPRs is essential for real-world human-robot collaboration in novel scenarios and this paper explores this gap between models and humans.

\end{document}